\documentclass[sigconf]{acmart}
\usepackage{algorithm}
\usepackage{algorithmic}
\usepackage{subcaption}
\usepackage{booktabs}
\usepackage{multirow}
\usepackage[most]{tcolorbox}
\AtBeginDocument{%
  }

\setcopyright{acmlicensed}
\copyrightyear{2025}
\acmYear{2025}
\acmDOI{XXXXXXX.XXXXXXX}
\acmConference[Conference acronym 'XX]{Make sure to enter the correct
  conference title from your rights confirmation email}{June 03--05,
  2018}{Woodstock, NY}
\acmISBN{978-1-4503-XXXX-X/2018/06}

\begin{document}

\title{GridMind: LLMs-Powered Agents for Power System Analysis and Operations}

%%
%% Author information - to be updated with actual authors
\author{Hongwei Jin\\Kibaek Kim}
\email{{jinh, kimk}@anl.gov}
% \orcid{0000-0000-0000-0000}
\affiliation{%
  \institution{Argonne National Laboratory}
  \department{Mathematics and Computer Science Division}
  \city{Lemont}
  \state{Illinois}
  \country{USA}
}

\author{Jonghwan Kwon}
\email{kwonj@anl.gov}
% \orcid{0000-0000-0000-0000}
\affiliation{%
  \institution{Argonne National Laboratory}
  \department{Energy Systems and Infrastructure Assessment Division}
  \city{Lemont}
  \state{Illinois}
  \country{USA}
}

% \renewcommand{\shortauthors}{Jin et al.}

\newcommand{\todo}[1]{{\color{red}{{\bf TODO: #1}}}}
\newcommand{\gridmind}{\textbf{GridMind} }
\newcommand{\savespace}{\vspace{0em}}
\begin{abstract}
The complexity of traditional power system analysis workflows presents significant barriers to efficient decision-making in modern electric grids. This paper presents GridMind, a multi-agent AI system that integrates Large Language Models (LLMs) with deterministic engineering solvers to enable conversational scientific computing for power system analysis. The system employs specialized agents coordinating AC Optimal Power Flow and N-1 contingency analysis through natural language interfaces while maintaining numerical precision via function calls. GridMind addresses workflow integration, knowledge accessibility, context preservation, and expert decision-support augmentation. Experimental evaluation on IEEE test cases demonstrates that the proposed agentic framework consistently delivers correct solutions across all tested language models, with smaller LLMs achieving comparable analytical accuracy with reduced computational latency. This work establishes agentic AI as a viable paradigm for scientific computing, demonstrating how conversational interfaces can enhance accessibility while preserving numerical rigor essential for critical engineering applications.
\end{abstract}




\received{20 February 2007}
\received[revised]{12 March 2009}
\received[accepted]{5 June 2009}

\maketitle
\savespace
\section{Introduction}

The emergence of Large Language Models (LLMs) and agentic AI systems is fundamentally transforming how we approach complex scientific and engineering problems. Recent advances in agentic AI—where autonomous agents can plan, execute, and reason about complex tasks—present unprecedented opportunities to revolutionize scientific and engineering computing workflows~\cite{schmidgall2025agent,brown2020language,openai2024gpt4}. Unlike simple conversational interfaces, agentic systems can orchestrate multiple computational processes, maintain context across complex analyses, and provide intelligent decision support through natural language interactions. A key capability of these agents is their ability to invoke existing computational functions and tools, ensuring that analytical requests are executed using reliable, validated methods and producing trustworthy results.

This paper examines the application of an agentic AI system in electric power systems, which requires complex planning and operation due to a wide range of technical and operational constraints. Modern power system analysis involves a diverse set of simulations across multiple time scales and levels of fidelity, including state-estimation, load and resource availability forecasting, alternative current optimal power flow (ACOPF), contingency analysis (CA), security-constrained unit commitment and economic dispatch, and long-term capacity planning with system reliability and resource adequacy assessments \cite{wu2019security,huneault2002survey,pagani2013power}. Each of these domains traditionally requires separate tools, programming expertise, and deep technical knowledge, leading to inefficient decision-making and fragmented workflows and analysis. 

This paper introduces \textbf{GridMind}, an early prototype of a \textbf{multi-agent} AI system that explores the potential of LLM-powered agents in power system analysis. Rather than relying on tool-driven workflows, our system demonstrates how conversation-based interactions can augment scientific computing, enabling domain experts to perform analyses more intuitively and efficiently.

% \todo{Figure : Conceptual diagram showing agentic workflow vs traditional tool-based approach}

\begin{figure}
    \centering
    \includegraphics[width=0.95\linewidth]{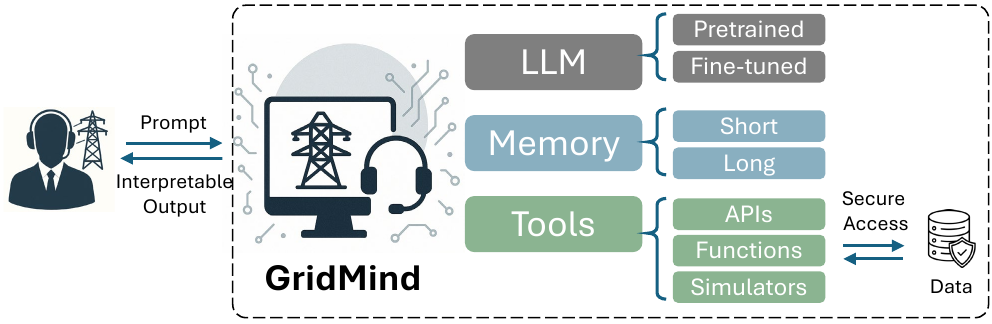}
    \caption{GridMind: LLM Agents for Powewr System}
    \label{fig:gridmind}
    \savespace
\end{figure}

\gridmind addresses four challenges in power system simulation and analysis: 
(1) \textbf{Workflow integration}—seamlessly connecting disparate analysis tasks through intelligent orchestration, 
(2) \textbf{Knowledge accessibility}—reducing programming barriers while maintaining technical rigor, 
(3) \textbf{Context preservation}—maintaining analytical coherence across complex multi-step processes, and 
(4) \textbf{Expert augmentation}—enhancing human decision-making through AI-powered insights and recommendations.
Our key contributions are threefold. First, we present a novel multi-agent architecture demonstrating how LLM agents orchestrate complex scientific engineering workflows through natural conversation. Second, we develop integrated power system analysis capabilities spanning multiple domains, coordinated through agentic workflows with conversational interfaces that interpret multi-step analytical requests. Third, we implement advanced context management preserving analytical state across agent interactions, with comprehensive evaluation confirming that agentic workflows maintain technical accuracy through structured function tools.

More specifically, \gridmind demonstrates how agentic AI lowers access barriers while supporting the natural iterative ``what-if'' analysis (e.g., adjust load levels, re-solve, inspect impacts). Automated tool selection and chaining compresses turnaround from question to vetted result. Built-in convergence, constraint, and data integrity checks keep numerical rigor visible in plain language instead of opaque solver logs. The system also serves as an instrumentation bench, logging solver metrics plus LLM backend latency, token usage, and occasional factual slips so reliability trends can be monitored. The prototype GridMind points toward similar gains in other scientific engineering domains that depend on multi-step analytical workflows.

\savespace
\section{Related Work}
% \todo{rephrase}

Agentic AI systems represent a fundamental shift from reactive to proactive AI, where agents can plan, execute, and reason about complex tasks autonomously~\cite{schmidgall2025agent}. Recent work has demonstrated the potential of LLM-powered agents in various domains, including software engineering~\cite{ding2023crosscodeeval}, scientific discovery~\cite{wang2023scientific}, and system administration~\cite{liu2023agentbench}. However, limited research exists on applying agentic workflows to scientific computing, particularly in engineering domains requiring rigorous numerical analysis such as power system analysis.
Moreover, multi-agent systems (MAS) have shown promise in coordinating complex computational tasks~\cite{wooldridge2009introduction,han2024llm}, but the integration of natural language processing with scientific computing agents remains largely unexplored. Our work contributes to this emerging field by demonstrating how conversational agents can orchestrate sophisticated engineering analyses.

The application of natural language processing to technical domains has gained significant attention with the advancement of large language models~\cite{li2023starcoder,nijkamp2022codegen}. Code generation and technical reasoning capabilities have enabled new paradigms for human-computer interaction in specialized domains~\cite{chen2021evaluating}. Frameworks such as LangChain~\cite{langchain2024}, PydanticAI~\cite{pydanticai2024}, and AutoGen~\cite{wu2023autogen} provide abstractions for tool invocation, memory/context handling, and (in AutoGen) multi-agent conversational patterns. PydanticAI explicitly enforces structured input/output validation via Pydantic models, whereas LangChain and AutoGen rely more on conventional Python typing plus runtime checks—so strong type safety and schema validation are not uniformly emphasized across all three. Despite these advances, deploying these general-purpose frameworks for tightly integrated, cross-domain scientific engineering workflows (e.g., coupling economic optimal power flow with large-scale contingency reliability assessment under a shared validated state) remains underexplored: current libraries supply generic orchestration primitives but limited built-in support for domain-specific numerical consistency checks, solver provenance tracking, and cross-analysis result fusion required in power system analytics.

The GridMind prototype focuses on two power system applications, ACOPF and CA, coordinated through an agentic workflow. ACOPF is a power system scheduling problem. It determines generator dispatch setpoints to minimize total operating costs while enforcing AC power flows and other technical and operational constraints~\cite{carpentier1962contribution,huneault2002survey}. Traditional optimization solution approaches include interior-point methods~\cite{quintana2000interior}, semidefinite programming relaxations~\cite{lavaei2011zero}, and metaheuristic algorithms~\cite{abido2002optimal}.

CA evaluates system reliability under T-1 outages of system assets. It determines post-contingency system operating conditions, including power flow limit violations, voltage violations, and involuntary load shedding~\cite{ejebe2007automatic}. CA also identifies critical system components whose failure could cause widespread violations~\cite{dobson2007complex}. Traditional CA tools require extensive setup and expertise~\cite{pavella2012transient,li2016real,sahraei2015real}.
Modern power system solver packages such as MATPOWER~\cite{zimmerman2011matpower}, PowerModels.jl~\cite{coffrin2018powermodels}, and PandaPower~\cite{thurner2018pandapower} provide robust implementations but require specialized programming knowledge and lack integrated workflow support across analysis domains.
\begin{figure}
    \centering
    \includegraphics[width=0.6\linewidth]{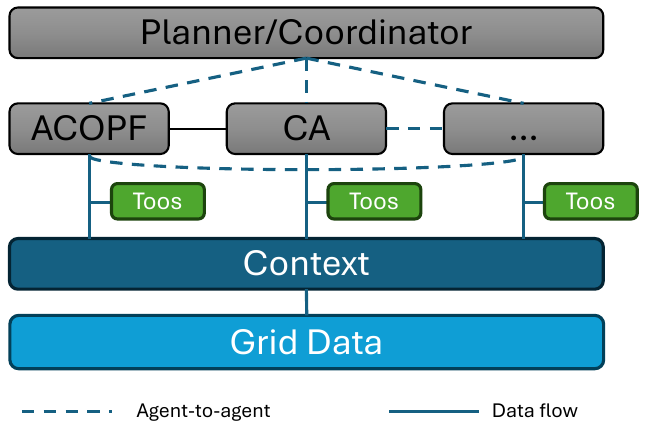}
    \caption{Multi-Agent Collaboration}
    \label{fig:multi-agent}
    \savespace
\end{figure}
Thus, the GridMind prototype aims to bridge these gaps by providing a user-friendly interface for complex power system analyses, enabling seamless collaboration between human experts and AI agents. It enables intuitive query formulation and result interpretation through natural language interactions and designs to benefit from the reasoning and thinking capabilities of LLMs for enhanced decision-making.
\savespace
\section{GridMind Agentic System Architecture}

\gridmind employs a multi-agent architecture that demonstrates how agentic AI can orchestrate complex scientific computing workflows. The core innovation of GridMind lies in our multi-agent orchestration approach (Figure~\ref{fig:multi-agent}), where specialized agents handle different aspects of power system analysis while maintaining coherent conversations and shared analytical context. The system comprises specialized agents working in coordination: (1) ACOPF agent, which specializes in economic scheduling of power systems and power flow analysis, (2) CA agent, which focuses on T-1 reliability assessment and critical element identification, (3) agent coordinator that manages inter-agent communication, context sharing, and complex multi-step analyses, and (4) planner agent analyzes user requests to determine appropriate agent assignment and workflow coordination. 

% This architecture enables sophisticated analytical workflows such as "solve IEEE 118 case, then run contingency analysis and identify the most critical elements for reinforcement."

% \todo{Add Figure: Multi-agent architecture diagram showing agent coordination and workflow orchestration}
\savespace
\subsection{Conversational Interface}

The interface is a thin front door: you type intent; the agents quietly parse it, plan a minimal sequence, call deterministic numerical solvers, check the numbers, and reply—no ad-hoc scripts. Shown in Appendix~\ref{sec:appendix-example}.

The agent handles intent and entity extraction (e.g., case id, buses, MW changes, outage scope) and sketches a compact plan. It then orchestrates typed tool calls for ACOPF, topology or load edits, and contingency sweeps, then layers validation: convergence flags, power balance tolerance, operating limits, and sanity checks on modified elements. A structured context keeps the latest solved state, applied diffs, and cached contingency fragments so only affected layers are recomputed. Every reported number is pulled from stored structured results, making the reply auditable and reproducible with timestamps and call metadata. New analytical tools can be registered with a schema; the planner notices capabilities without refactoring core logic. The result is a deterministic loop—parse, plan, invoke, validate, narrate, persist—that preserves engineering rigor while trimming the friction of multi-step exploratory studies.
\savespace
\subsection{Domain Agents}

\subsubsection{ACOPF Agent}

The ACOPF agent is implemented as an LLM-driven computational assistant that fuses four pillars: (1) a pretrained reasoning model (i.e., the LLM), (2) structured conversational memory to maintain context state, (3) a vetted toolbox of deterministic power system solvers, and (4) carefully engineered prompts that bind user intent to verifiable analytical actions. Rather than hallucinating numerical outcomes, the agent plans in language, invokes trusted functions for every quantitative step, then interprets the returned structured results back into domain-aware explanations. And it is designed as a ``reason-act-reflect'' loop in which the agent can reason about the problem, act on it, and reflect on the results to ensure correctness and reliability.

\textbf{LLM core and prompting.} The language model supplies chain-of-thought style reasoning: it decomposes a user request (e.g., ``Evaluate the economic impact of removing transmission line (Line) between buses 37 and 40 in the IEEE 118-bus case'') into actionable subgoals: (a) load or confirm the target network, (b) verify whether a base ACOPF solution exists in context, (c) schedule a re-dispatch under the modified topology, (d) compare objective cost, constraint margins, and voltage profile, (e) prepare a concise, auditable report. System prompts constrain behavior (``Never fabricate solver outputs; always call tools for numerical data'') while tool-specific prompts provide schemas for input/output validation.

\textbf{Memory (context).} A structured in-session memory object stores: current case metadata (case id, bus/gen counts, last solve timestamp), the latest feasible dispatch, contingency results cache, and any modifications (e.g., load adjustments, line outages) as a chronological diff log. Before acting, the agent replays relevant slices of this state to ground its reasoning (``A solved ACOPF for IEEE 57 exists with generation cost \$41{,}532.17; two loads were increased previously.''). This prevents redundant solves and enables cumulative what-if analysis.

\textbf{Tool integration.} Every numerical claim originates from registered tools, e.g., PandaPower; 
% for example:
% \begin{itemize}
%   \item \texttt{solve\_acopf(case, options)} $\rightarrow$ structured \texttt{ACOPFSolution}
%   \item \texttt{apply\_load\_change(case, bus, mw\_delta)} $\rightarrow$ updated network snapshot
%   \item \texttt{compute\_line\_flows(case)} $\rightarrow$ per-branch magnitudes / ratings
% \end{itemize}
The agent selects tools by matching inferred subtask types (``needs re-optimization'') with tool capability descriptors, then emits a JSON-typed call specification. Post-execution, returned objects are validated (e.g., convergence flag, max power balance mismatch $< 10^{-4}$ p.u.) before interpretation. If validation fails, the agent triggers an automatic recovery path (adjust solver tolerances, fall back to an alternative algorithm, or request clarification).

\textbf{Trust and auditability.} Every numeric in the narrative maps to a field in a stored tool output object (from structured data). This design ensures that the agent does not fabricate numbers or make unverifiable claims. Instead, it provides a transparent audit trail: each request will be traceable to its originating data sources and use proper tools to solve the case in a numerical way, instead of completely relying on LLMs' reasoning.

In sum, the single ACOPF agent exhibits disciplined cognitive cycles—grounded planning, tool-driven action, reflective validation, and evidence-based explanation—yielding trustworthy, inspectable power system analytics through natural language interaction.

\subsubsection{Contingency Analysis Agent}
% The contingency analysis agent acts as a reliability strategist: once a stable base operating point is available, it systematically imagines the grid with each transmission element removed (T-1) and asks whether the system can still operate within its thermal and voltage envelopes. 
% Rather than dumping raw overload tables, it runs a disciplined loop—prepare the modified topology, invoke trusted power flow tools (e.g., \textit{runpp} in pandapower), validate numerical solution, then interpret the pattern of stress in the context of prior runs. 
% It keeps a rolling memory of recurring bottlenecks, voltage weakness pockets, and the severity of post-contingency violations that would require corrective actions, allowing it to distinguish a benign single overload from an outage that simultaneously creates multiple thermal and voltage violations. Its language reasoning layer never fabricates numbers; every cited percentage or voltage dip is anchored to structured solver outputs cached with provenance. When several contingencies compete for ``most critical,'' it narrates trade-offs—compound thermal congestion versus emergent low-voltage risk versus potential load shedding requirements—producing an auditable story rather than an opaque score. This design turns brute-force T-1 enumeration into an interactive diagnostic dialogue, surfacing not only which elements are risky but why they matter and what mitigation strategies—reactive support, targeted reinforcement, topology adjustments, or operational procedures—could most efficiently restore security margins.

The contingency analysis agent systematically evaluates grid reliability by simulating T-1 outages for each transmission element. It executes a structured workflow: modify topology, invoke power flow solvers (e.g., \textit{runpp} in pandapower), validate solutions, and interpret stress patterns. The agent maintains memory of recurring bottlenecks and voltage vulnerabilities, distinguishing between minor single overloads and outages causing multiple simultaneous violations. Rather than fabricating numbers, it anchors every cited metric to structured solver outputs with full provenance. When ranking competing contingencies, it provides auditable narratives comparing thermal congestion, voltage risks, and potential load shedding requirements. This transforms brute-force T-1 enumeration into interactive diagnostics that identify critical elements and suggest targeted mitigation strategies.

\subsubsection{Critical Element Identification}
The contingency analysis agent uses LLM reasoning to synthesize structured solver outputs (line loading percentages, voltage deviations, load curtailment) into ranked critical transmission elements. Rather than using single metrics, the LLM analyzes evidence from tools: clusters of thermal overloads exceeding 110-115\% ratings, voltage excursions below 0.94 p.u., and redispatch triggering load shedding. For instance, removing Line 54-59 in IEEE 118 forcing two lines to 118\% and 121\% loading while dropping a bus to 0.928 p.u. elevates criticality due to combined thermal and voltage stress.
The agent transparently articulates comparisons: ``\textit{Outage A causes three overloads requiring 12 MW curtailment, while Outage B causes one marginal overload—therefore A ranks higher.}'' This generates actionable recommendations for capacity reinforcement, reactive support, or remedial switching. By grounding ranking steps in validated numerical context, the LLM avoids hallucination and produces auditable justifications supporting operational decision-making.

\noindent\textbf{Abridged dialogue showcase with steps in agents. Please check Appendix~\ref{sec:appendix-example} for complete examples.}
\begin{verbatim}
User : Solve IEEE 118.
Agent: 1. (understand the case to be solved) -> reasoning
       2. (extract relevant parameters)      -> reasoning
       3. (plan solution strategy)           -> reasoning
       4. (invoke ACOPF solver)         -> function tools
       5. (validate results)            -> function tools
       6. (narrate findings)                  -> response
User: Increase the load for bus 10 to 50MW
Agent: 1. (understand the task to solve)     -> reasoning
       2. (retrieve current net status)        -> context
       3. (prepare data for tools)           -> reasoning
       4. (invoke ACOPF solver again)   -> function tools
       5. (validate results)            -> function tools
       6. (summary)                           -> response
User: what's the most critical contingencies in this network
Agent: 1. (understand task)                  -> reasoning
       2. (shift from ACOPF agent to AC agent) -> shared context
       3. (run contingency analysis)    -> function tools
       4. ...
\end{verbatim}
\savespace
\subsection{Agent Context Management}
\paragraph{Data Models and Type Safety}
\gridmind relies on a set of strongly typed, structured data models to keep every agent grounded in the same verifiable representation of the power system and its analytical state. Instead of passing loosely defined dictionaries, the system serializes and validates each object—network snapshots, optimization solutions, contingency outcomes, and shared conversational context—against explicit schemas. 
A unified \texttt{PowerSystem} model captures buses, generators, branches, transformers, and metadata;
an \texttt{ACOPFSolution} object stores total cost, dispatch, losses, constraint margins, and convergence flags; 
a \texttt{ContingencyAnalysisResult} aggregates per-outage thermal and voltage impacts plus criticality narratives; 
an \texttt{AgentContext} records the active case, cached solutions, applied modifications, and provenance; 
and a \texttt{WorkflowState} traces multi-step analytical plans and their completion status. 
These structured types do more than enforce shape—they provide semantic anchors the LLM can reference when planning actions, summarizing results, or deciding whether prior computations can be reused. 
By surfacing field names (e.g., total\_cost, max\_thermal\_loading, min\_voltage\_pu) directly in tool outputs, the model maps natural language intents (``compare post-contingency voltage minima'') to exact data fields, reducing ambiguity and suppressing hallucinated attributes. 
Type validation also acts as a safety net: malformed or incomplete tool returns trigger automatic recovery paths instead of silently corrupting downstream reasoning. 
In practice, this schema layer converts free-form dialogue into a controlled loop of intent, structured call, validated result, grounded explanation.
\savespace
\subsection{Cross-Agent Context Management}
Agents collaborate through a single structured, versioned session state capturing: (a) the active network plus incremental diffs (load shifts, outages, parameter edits), (b) validated numerical artifacts (latest feasible ACOPF solution, cached power-flow / per-contingency snapshots), and (c) provenance (solver options, timestamps, tool versions, validation flags). After a solve, the ACOPF agent deposits a typed \texttt{ACOPFSolution} (dispatch, objective cost, losses, voltage extrema, branch loadings, constraint margins) instead of only prose. The CA agent inspects freshness against the diff log to decide whether it can reuse that base point or must trigger a selective re-solve.

Cross-agent transfer is strictly schema-bound (\texttt{ACOPFSolution}, \texttt{ContingencyResultSet}) so planning references concrete fields (e.g., base\_objective\_cost, min\_voltage\_pu, max\_thermal\_loading) and avoids hallucination. Each outage evaluation is cached under a composite key (case + outage + diff hash); criticality ranking streams those cached records to detect recurring overload corridors or voltage depressions and writes back ranked justifications as structured data plus a derived human summary. A normalized change log appends every modification; agents replay only relevant diffs to reconstruct required state. Session persistence serializes baseline, diffs, artifacts, contingency cache, and rankings for seamless resumption. This disciplined produce-validate-consume loop lets compound requests (``solve, assess T-1 risk, rank reinforcements'') execute efficiently, coherently, and reproducibly.
\savespace
\section{Experimental Evaluation}

We conducted comprehensive experiments to evaluate \gridmind's agentic capabilities across multiple dimensions: technical accuracy for domain tasks, efficiency of agents by evaluating the response time from LLMs, and robustness under various input perturbations.
We implemented the agentic workflow using PydanticAI~\cite{pydanticai2024}, and systematically evaluate the agentic performance based on a set of LLMs including OpenAI's GPT-5, GPT-5-mini, GPT-5-nano, GPT-o3, GPT-o4-mini, and Anthropic's Claude 4 Sonnet. The backend power system solvers are implemented using PandaPower~\cite{thurner2018pandapower}. We deployed the system on a local machine with 32-core Intel Xeon Gold 5317 CPU, 256GB RAM. The LLMs are accessed through either remote APIs (e.g., OpenAI, Anthropic, proxy server Argo).
Our evaluation employed standard IEEE test cases representing different system scales and complexity levels, IEEE 14, 30, 118, and 300-bus systems~\cite{pstca1999}.
\savespace
\begin{figure*}[ht]
    \centering
    \includegraphics[width=\linewidth]{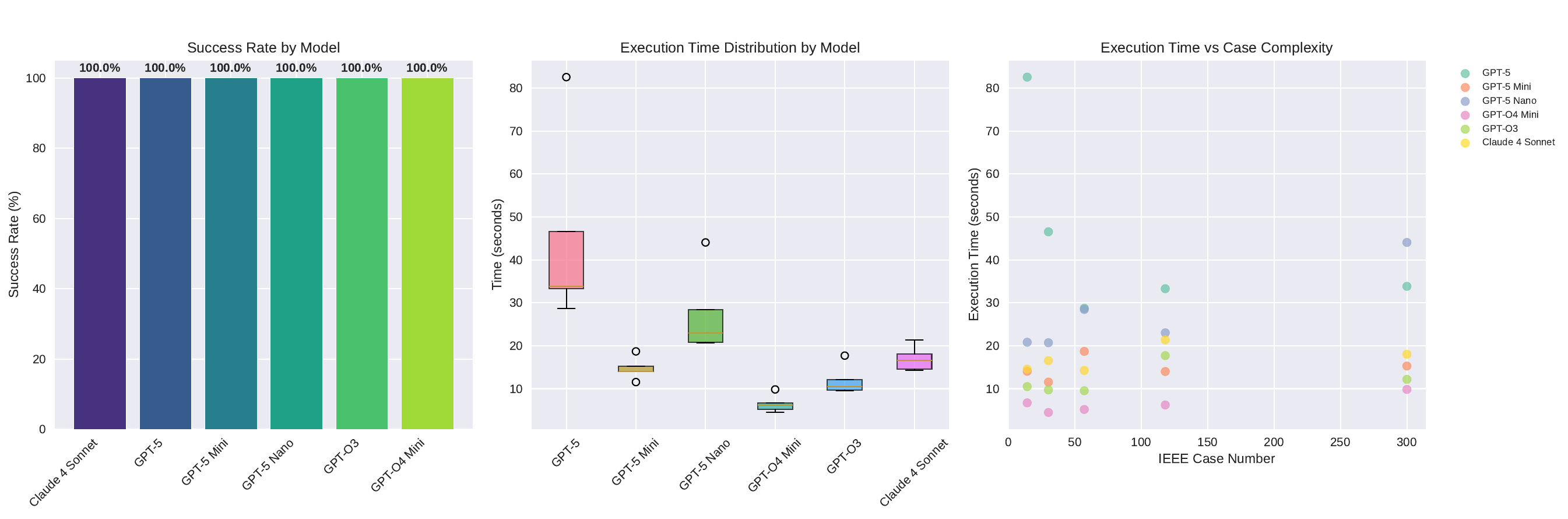}
    \caption{Agent performance in terms of accuracy and execution time}
    \label{fig:acopf_overview}
    \savespace
\end{figure*}
\subsection{ACOPF Agent}
The evaluation of \gridmind's ACOPF agent demonstrates remarkable consistency in solution quality across different language models while revealing significant variations in computational efficiency. Figure~\ref{fig:acopf_overview} presents a comprehensive performance overview with three key metrics: success rates, execution time distributions, and execution time versus case complexity.

The left figure shows that all tested models achieve 100\% success rates across IEEE test case118, demonstrating that function-calling capabilities enable reliable technical accuracy regardless of the underlying LLM. This consistent performance validates our hypothesis that deterministic solver invocation through structured tool calls maintains numerical rigor even with smaller language models.

The middle figure reveals substantial differences in execution time distributions across models with case118 by running 5 times. GPT-o4-mini exhibits the most variable performance with execution times less than 10 seconds. In contrast, most recent GPT-5, GPT-5-mini, GPT-5 Nano and Claude 4 Sonnet demonstrate more time to reason, potentially due to their large models and complex reasoning processes. This is interesting to note that with function tool capabilities, there is a potential trade-off between model size and reasoning speed, as smaller models can achieve comparable accuracy with lower latency.

The right figure illustrates the relationship between system complexity (measured by IEEE case number as a proxy for network size) and execution time. With the increase of case size, there is no significant trends of time to get the solution, indicating that the agentic framework effectively manages complexity without incurring additional computational costs from LLMs. However, calling the tools for specific case is still subject to the case size and complexity of its original problem, e.g., ACOPF.

These findings highlight that agentic scientific computing can achieve both technical rigor and computational economy by leveraging function-calling architectures that combine efficient language models for workflow orchestration with validated domain-specific tools for numerical calculations.

\savespace
\subsection{T-1 Contingency Analysis Agent}
Unlike the ACOPF agent that can compare the solution directly from a reference solver, the CA agent cannot retrieve a ground solution for the problem. CA, as a result, relies on the LLM's ability to reason about potential contingencies and their impacts on system reliability. CA agent will first calculate the power flow for the base case, without any contingencies, and then iteratively apply each identified contingency to assess its impact on system reliability.
Therefore, we conducted a series of experiments based on different LLMs to identify the top-5 most critical lines, and let agent applies tools (power flow solver) to calculate the maximum overload percentage for each contingency. The results are summarized in Table~\ref{tab:exp_ca_agent}.
\begin{table}[]
    \centering
    \caption{CA Agent Performance (case118)}
    \label{tab:exp_ca_agent}
    \begin{tabular}{r|ccc}
    \hline
        Model & Time (s) & Critical Lines (idx) & Max Overload \% \\
        \hline
        GPT-5 & 92.7 & 6, 7, 0, 171, 49 & 137 \\
        GPT-5 Mini & 24.8 & 7, 0, 171, 49, 9 & 165 \\
        GPT-5 Nano & 26.2 & 6, 7, 0, 171, 49 & 137 \\
        GPT-o4 Mini & 34.2 & 6, 7, 0, 171, 49 & 137 \\
        GPT-o3 & 24.6 & 6, 7, 0, 171, 49 & 137 \\
        Claude 4 Sonnet & 63.3 & 6, 7, 0, 171, 49 & 137\\
        \hline
    \end{tabular}
\end{table}
The results show significant variation in computational efficiency, with GPT-5 Mini and GPT-o3 achieving the fastest execution times at approximately 24-25 seconds, while GPT-5 required the longest processing time at 92.7 seconds. Interestingly, most models identified the same set of critical transmission lines (indices 6, 7, 0, 171, 49) and achieved identical maximum overload percentages of 137\%, suggesting consistent analytical accuracy across different AI architectures. The notable exception is GPT-5 Mini, which identified a slightly different set of critical lines (replacing index 6 with index 9) and reported a higher maximum overload of 165\%, indicating either a different analytical approach or potentially identifying additional stress conditions in the power system that other models may have missed. This is mainly due to the architectural and training differences from various LLMs and their build-in reasoning nature.

% \subsection{Computational Performance and Scalability}

% \todo{examine the cost of different LLMs in terms of computational resources and efficiency, available models from Argo (remote) or Ollama (local), think about metrics to be collected (e.g., latency, throughput, resource utilization, tokens, etc.)}

% Performance analysis demonstrates excellent scalability characteristics:

% \todo{Figure: Performance scaling curves for ACOPF and contingency analysis across system sizes}

% \subsection{Comparative Analysis with Traditional Tools}

% Direct comparison with traditional power system analysis tools reveals significant advantages:

% \todo{Table: Comparison of workflow efficiency and user experience metrics}

\textbf{Discussion.} \gridmind demonstrates how agentic AI can transform scientific computing by converting fragmented workflows into fluent conversational interfaces. Our prototype shows that LLMs, when bound to deterministic tools and structured state management, can orchestrate multi-step analyses while maintaining numerical rigor.

\paragraph{Addressing LLM Hallucination} A critical concern with LLM-based agents is hallucination—generating plausible but incorrect information. \gridmind mitigates this through: (1) structured function calls that prevent numerical fabrication, (2) rigorous data validation using Pydantic schemas, (3) grounding all quantitative claims in solver outputs, and (4) comprehensive result verification before response generation. This architecture ensures that while agents may hallucinate in reasoning steps, all reported numerical results originate from validated computational tools.

Future extensions include expanding to adjacent domains, implementing richer coordination protocols, and developing human-AI interaction policies for critical engineering applications. These results position agentic frameworks as organizing principles for scientific computing ecosystems that harmonize expressiveness, rigor, and accessibility.
% \savespace
\section{Conclusion}

This paper introduced \gridmind, a groundbreaking multi-agent AI system that demonstrates the transformative potential of agentic workflows in scientific engineering. Through the seamless integration of specialized agents for ACOPF optimization and T-1 contingency analysis, our system represents a paradigm shift from tool-based to conversational AI-driven scientific computing.
Our key contributions demonstrate that agentic AI represents a viable and transformative paradigm for complex scientific workflows. The multi-agent architecture successfully coordinates sophisticated power system analyses through natural language interfaces while preserving the numerical precision essential for engineering applications. Notably, our evaluation reveals that the choice of underlying LLM significantly influences both system performance and reliability. Contrary to expectations, smaller language models can achieve competitive analytical accuracy with substantially reduced reasoning latency, suggesting that the most advanced model is not necessarily optimal for all agentic scientific computing tasks.

\begin{acks}
  This material is based upon work supported by Laboratory Directed Research and Development (LDRD) funding from Argonne National Laboratory, provided by the Director, Office of Science, of the U.S. Department of Energy under contract DE-AC02-06CH11357. This research used resources of the Argonne Leadership Computing Facility at Argonne National Laboratory, which is supported by the Office of Science of the U.S. Department of Energy under contract DE-AC02-06CH11357.
\end{acks}

% \section{Appendices}

% \begin{acks}
% \end{acks}
% \clearpage
% \newpage
\bibliographystyle{ACM-Reference-Format}
\bibliography{ref.bib}

\appendix

\section{Case Studies}
\begin{table}
  \centering
  \caption{Test cases}
  \label{tab:ieee_cases}
  \begin{tabular}{c|ccccc}
    \hline
    Case & Bus & Gen & Load & AC line & Transformers \\
    \hline
    IEEE 14 & 14 & 5 & 11 & 17 & 3 \\
    IEEE 30 & 30 & 6 & 21 & 41 & 4 \\
    IEEE 57 & 57 & 7 & 42 & 63 & 17 \\
    IEEE 118 & 118 & 54 & 99 & 175 & 11 \\
    IEEE 300 & 300 & 68 & 193 & 283 & 128 \\
    \hline
  \end{tabular}
\end{table}
\section{Implementation Details}

\subsection{System Requirements}
\begin{itemize}
  \item Python 3.10 or higher
  \item PydanticAI (version 0.7 or higher) framework for agents, function tools, and context management
  \item PandaPower for power system simulation and analysis
  \item NumPy and SciPy for numerical computations
  \item Rich library for CLI interface formatting
  \item PandaPower (runpp for powerflow) for line contingencies.
\end{itemize}

\subsection{Supported IEEE Test Cases}
Table~\ref{tab:ieee_cases} summarizes the supported IEEE test cases, highlighting their key characteristics and complexities.

\subsection{Construct Agent}

\subsubsection{ACOPF Agent}
\paragraph{System prompt} Shown in Figure~\ref{fig:acopf-prompt}
\begin{figure*}[ht]
\begin{lstlisting}[basicstyle=\footnotesize\ttfamily, breaklines=true]
You are an expert ACOPF (AC Optimal Power Flow) agent for power system analysis.

Your capabilities include:
1. Solving ACOPF problems for standard IEEE test cases (14, 30, 57, 118, 300 bus systems)
2. Modifying system parameters (loads, generation limits, etc.) and re-solving
3. Validating solutions by checking power flows, voltage limits, and line loadings
4. Assessing solution quality and providing recommendations
5. Engaging in conversational interactions about power system optimization

You have access to the following tools:
- solve_acopf_case: Load and solve an IEEE test case
- modify_bus_load: Modify load at a specific bus and re-solve
- get_network_status: Get current network and solution status

When users ask to solve a case, use the solve_acopf_case tool with the case name.
When users ask to modify loads, use the modify_bus_load tool with the specified parameters.
When users ask about current status, use the get_network_status tool.

Always provide clear explanations of results, including objective values and any constraint violations.
Be professional, accurate, and educational in your responses.
\end{lstlisting}
\caption{ACOPF Agent System Prompt}
\label{fig:acopf-prompt}
\end{figure*}

\paragraph{Function Tools}
\begin{verbatim}
ACOPF Agent: 
    solve_acopf_case
    modify_bus_load
    get_network_status
\end{verbatim}

\subsubsection{Contingency Analysis Agent}
\paragraph{System Prompt} Shown in Figure~\ref{fig:contingency-prompt}
\begin{figure*}[ht]
\begin{lstlisting}[basicstyle=\footnotesize\ttfamily, breaklines=true]
You are an expert Contingency Analysis agent for power system reliability assessment.

Your capabilities include:
1. Solving base case ACOPF problems for standard IEEE test cases
2. Running comprehensive N-1 contingency analysis
3. Analyzing specific contingencies (line outages, transformer outages)
4. Identifying critical contingencies and system vulnerabilities
5. Assessing voltage violations and equipment overloads
6. Providing recommendations for system reinforcement

You have access to the following tools:
- solve_base_case: Load and solve base case before contingency analysis
- run_n1_contingency_analysis: Run comprehensive N-1 analysis
- analyze_specific_contingency: Analyze a specific element outage
- get_contingency_status: Get current analysis status and results

When users ask to analyze contingencies, first ensure a base case is solved, then run the appropriate analysis.
Always provide clear explanations of critical contingencies, violations, and recommendations.
Be professional, accurate, and focus on system reliability and security.
\end{lstlisting}
\caption{Contingency Analysis Agent System Prompt}
\label{fig:contingency-prompt}
\end{figure*}

\paragraph{Function Tools}
\begin{verbatim}
CA Agent:
    solve_base_case
    run_n1_contingency_analysis
    analyze_specific_contingency
    get_contingency_status
\end{verbatim}

\subsection{Multi-Agent Workflow Capabilities}

The system supports sophisticated analytical sequences including:
\begin{itemize}
  \item Sequential analysis workflows (ACOPF followed by contingency analysis)
  \item Comparative studies (economic vs. security-constrained operation)
  \item Sensitivity analysis (parameter modifications with impact assessment)
  \item Cross-domain result correlation and insight generation
\end{itemize}

% \todo{Add detailed API reference and code examples for agent interaction patterns}

\section{Data schema}
\noindent\textbf{Illustrative schema fragment:}
\begin{verbatim}
class ACOPFSolution(BaseModel):
    case_name: str
    solved: bool
    objective_cost: float
    gen_dispatch_mw: dict[str, float]
    branch_loading: list[BranchLoading]
    min_voltage_pu: float
    max_voltage_pu: float
    convergence_message: str
    
class SolutionQuality(BaseModel):
    overall_score: float = Field(ge=0.0, le=10.0)
    convergence_quality: float = Field(ge=0.0, le=10.0)
    constraint_satisfaction: float = Field(ge=0.0, le=10.0)
    economic_efficiency: float = Field(ge=0.0, le=10.0)
    system_security: float = Field(ge=0.0, le=10.0)
    detailed_metrics: Dict[str, Any] = Field(default_factory=dict)
    recommendations: List[str] = Field(default_factory=list)
\end{verbatim}

\section{Example Agentic Workflows}
\label{sec:appendix-example}

\subsection{CLI interface}
\begin{figure}[ht]
  \centering
  \includegraphics[width=0.9\linewidth]{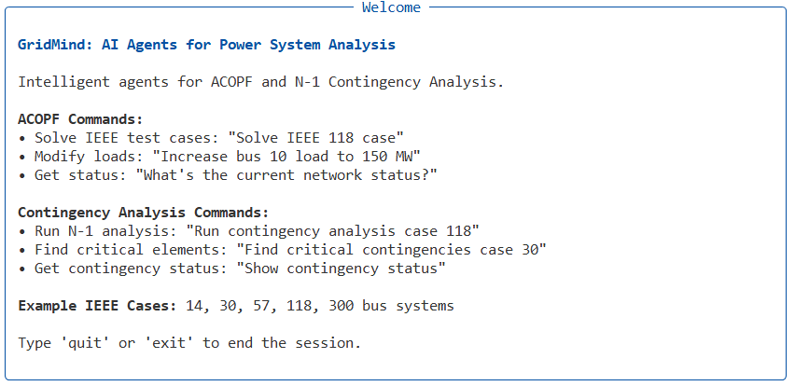}
  \caption{Conversational interface (CLI prototype) driving multi-agent analysis.}
  \label{fig:interface}
\end{figure}

\subsection{Single-Domain Analysis Examples}

\subsubsection{ACOPF Agent}
\begin{figure*}
    \centering
    \includegraphics[width=1\linewidth]{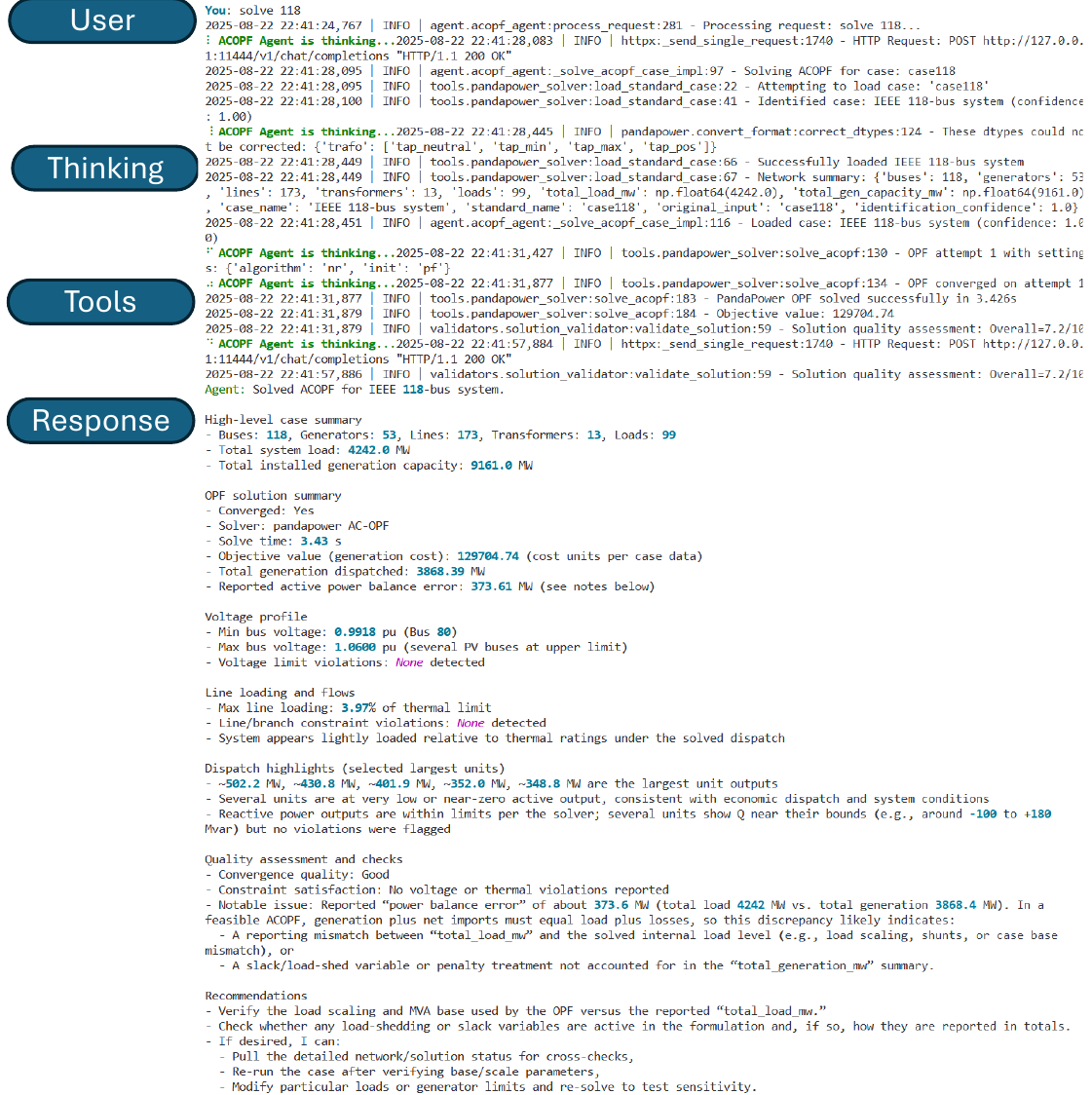}
    \caption{ACOPF Agent Workflow}
    \label{fig:acopf_agent_workflow}
\end{figure*}

Figure~\ref{fig:acopf_agent_workflow} shows the ACOPF agent for solving optimal power flow problems.

\subsubsection{Contingency Analysis}
\begin{figure*}
    \centering
    \includegraphics[width=1\linewidth]{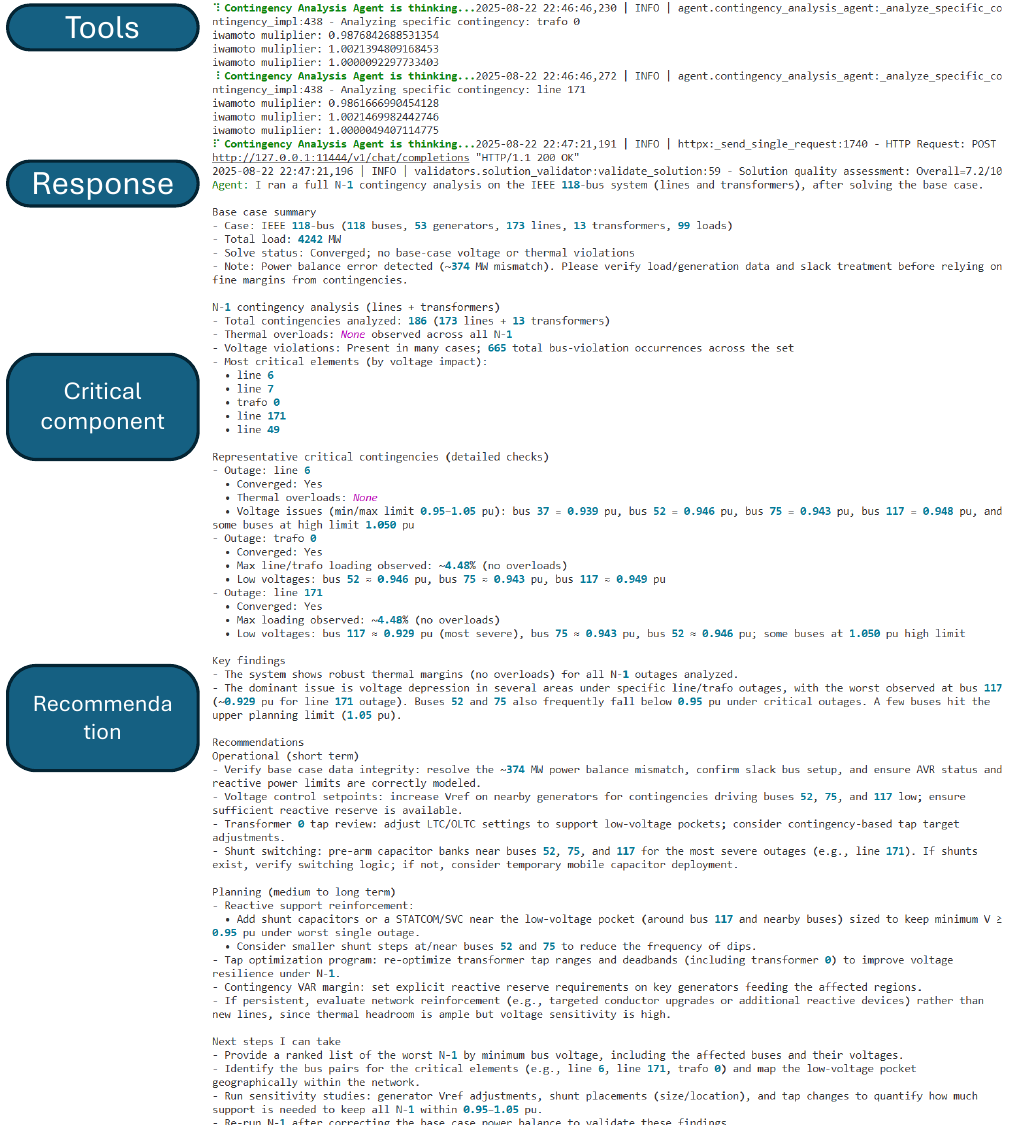}
    \caption{Contingency Analysis Agent Workflow}
    \label{fig:contingency_analysis_agent_workflow}
\end{figure*}

Figure~\ref{fig:contingency_analysis_agent_workflow} shows the contingency analysis agent for T-1 reliability assessment.

\subsection{Multi-Agent Workflow Examples}

\subsubsection{Cross-Domain Analysis Workflow}
\begin{figure*}
    \centering
    \includegraphics[width=\linewidth]{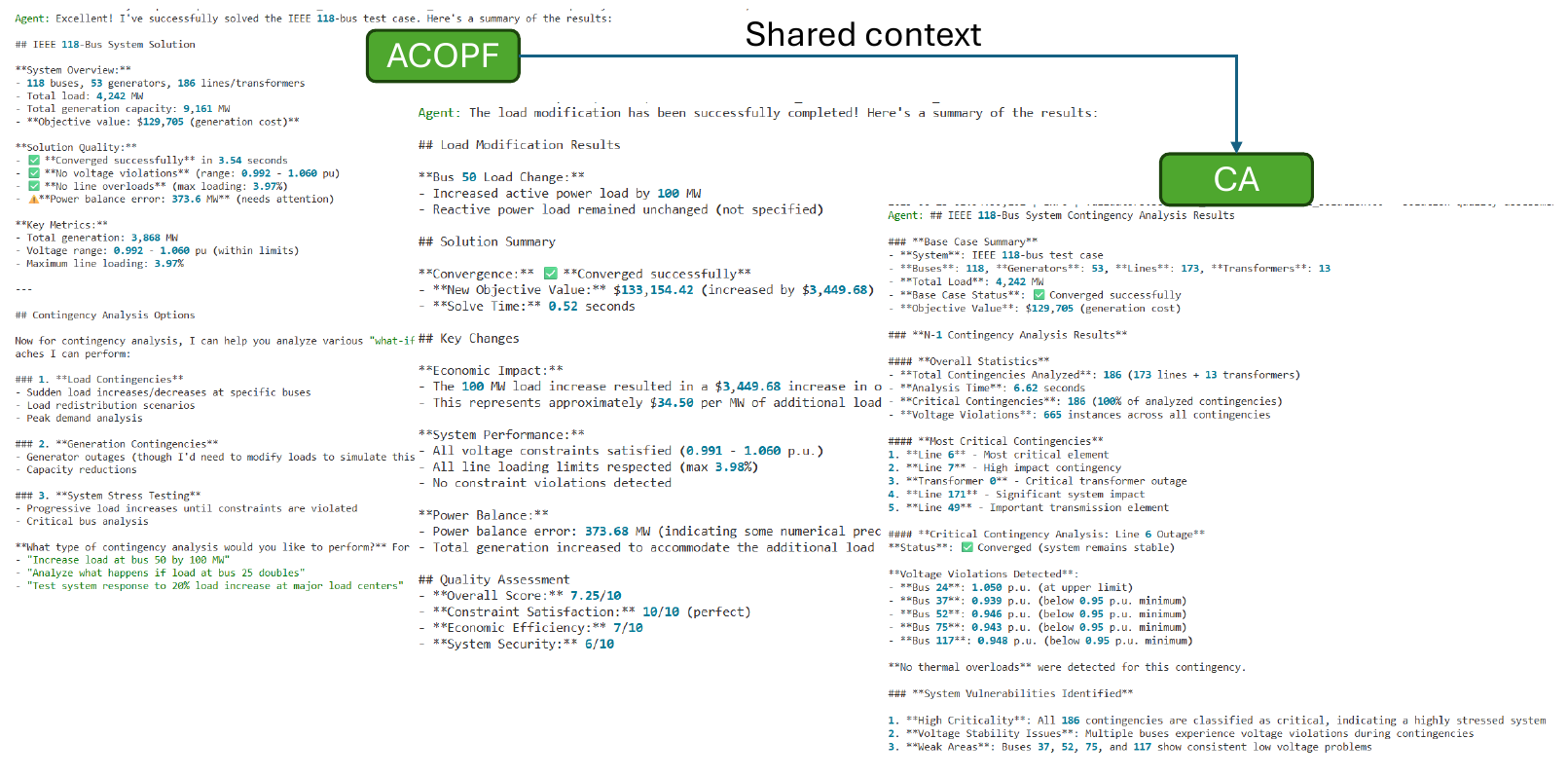}
    \caption{Multi-Agent Workflow Example}
    \label{fig:multi-agent-demo}
\end{figure*}

Figure~\ref{fig:multi-agent-demo} shows the agentic workflow, particularly from ACOPF agent to CA agent with shared context.
\begin{verbatim}
User: Solve IEEE 118 case, then run contingency analysis 
      and identify critical elements for reinforcement
\end{verbatim}

\end{document}